\PassOptionsToPackage{unicode}{hyperref}
\PassOptionsToPackage{hyphens}{url}
\documentclass[
  11pt,
]{article}
\usepackage{lmodern}
\usepackage{amssymb,amsmath}
\usepackage{ifxetex,ifluatex}
\ifnum 0\ifxetex 1\fi\ifluatex 1\fi=0 % if pdftex
  \usepackage[T1]{fontenc}
  \usepackage[utf8]{inputenc}
  \usepackage{textcomp} % provide euro and other symbols
\else % if luatex or xetex
  \usepackage{unicode-math}
  \defaultfontfeatures{Scale=MatchLowercase}
  \defaultfontfeatures[\rmfamily]{Ligatures=TeX,Scale=1}
\fi
\IfFileExists{upquote.sty}{\usepackage{upquote}}{}
\IfFileExists{microtype.sty}{% use microtype if available
  \usepackage[]{microtype}
  \UseMicrotypeSet[protrusion]{basicmath} % disable protrusion for tt fonts
}{}
\makeatletter
\@ifundefined{KOMAClassName}{% if non-KOMA class
  \IfFileExists{parskip.sty}{%
    \usepackage{parskip}
  }{% else
    \setlength{\parindent}{0pt}
    \setlength{\parskip}{6pt plus 2pt minus 1pt}}
}{% if KOMA class
  \KOMAoptions{parskip=half}}
\makeatother
\usepackage{xcolor}
\IfFileExists{xurl.sty}{\usepackage{xurl}}{} % add URL line breaks if available
\IfFileExists{bookmark.sty}{\usepackage{bookmark}}{\usepackage{hyperref}}
\hypersetup{
  hidelinks,
  pdfcreator={LaTeX via pandoc}}
\usepackage[margin=1in]{geometry}
\usepackage{longtable,booktabs}
\usepackage{etoolbox}
\makeatletter
\patchcmd\longtable{\par}{\if@noskipsec\mbox{}\fi\par}{}{}
\makeatother
\IfFileExists{footnotehyper.sty}{\usepackage{footnotehyper}}{\usepackage{footnote}}
\makesavenoteenv{longtable}
\usepackage{graphicx}
\makeatletter
\def\maxwidth{\ifdim\Gin@nat@width>\linewidth\linewidth\else\Gin@nat@width\fi}
\def\maxheight{\ifdim\Gin@nat@height>\textheight\textheight\else\Gin@nat@height\fi}
\makeatother
\setkeys{Gin}{width=\maxwidth,height=\maxheight,keepaspectratio}
\makeatletter
\def\fps@figure{htbp}
\makeatother
\providecommand{\tightlist}{%
  \setlength{\itemsep}{0pt}\setlength{\parskip}{0pt}}
\usepackage{graphicx}
\usepackage{float}
\usepackage{xurl}
\usepackage{newunicodechar}
\usepackage{amssymb}
\usepackage{textcomp}

\usepackage{fvextra}
\fvset{breaklines=true, breakanywhere=true, breaksymbol={}}
\RecustomVerbatimEnvironment{verbatim}{Verbatim}{breaklines=true,breakanywhere=true,breaksymbol={}}

\usepackage{etoolbox}
\AtBeginEnvironment{longtable}{\small}

\usepackage{iftex}

\ifPDFTeX

\newunicodechar{ℝ}{\ensuremath{\mathbb{R}}}

\newunicodechar{⁶}{\textsuperscript{6}}
\newunicodechar{ⁿ}{\textsuperscript{n}}
\newunicodechar{ᵀ}{\textsuperscript{T}}

\newunicodechar{₀}{\textsubscript{0}}
\newunicodechar{₁}{\textsubscript{1}}
\newunicodechar{₂}{\textsubscript{2}}
\newunicodechar{₃}{\textsubscript{3}}
\newunicodechar{₄}{\textsubscript{4}}
\newunicodechar{₅}{\textsubscript{5}}
\newunicodechar{₆}{\textsubscript{6}}
\newunicodechar{₇}{\textsubscript{7}}
\newunicodechar{₈}{\textsubscript{8}}
\newunicodechar{₉}{\textsubscript{9}}

\newunicodechar{ₙ}{\textsubscript{n}}
\newunicodechar{ₖ}{\textsubscript{k}}
\newunicodechar{ₜ}{\textsubscript{t}}
\newunicodechar{ᵢ}{\textsubscript{i}}
\newunicodechar{ⱼ}{\textsubscript{j}}

\newunicodechar{₊}{\textsubscript{+}}
\newunicodechar{₋}{\textsubscript{-}}

\newunicodechar{−}{\ensuremath{-}}
\newunicodechar{∈}{\ensuremath{\in}}
\newunicodechar{∉}{\ensuremath{\notin}}
\newunicodechar{≤}{\ensuremath{\leq}}
\newunicodechar{≥}{\ensuremath{\geq}}
\newunicodechar{≠}{\ensuremath{\neq}}
\newunicodechar{≈}{\ensuremath{\approx}}
\newunicodechar{→}{\ensuremath{\rightarrow}}
\newunicodechar{←}{\ensuremath{\leftarrow}}
\newunicodechar{×}{\ensuremath{\times}}
\newunicodechar{∑}{\ensuremath{\sum}}
\newunicodechar{∞}{\ensuremath{\infty}}

\newunicodechar{ⓘ}{\textcircled{\scriptsize i}}

\newunicodechar{θ}{\ensuremath{\theta}}
\newunicodechar{μ}{\ensuremath{\mu}}
\newunicodechar{σ}{\ensuremath{\sigma}}
\newunicodechar{τ}{\ensuremath{\tau}}
\newunicodechar{φ}{\ensuremath{\varphi}}
\newunicodechar{α}{\ensuremath{\alpha}}
\newunicodechar{β}{\ensuremath{\beta}}
\newunicodechar{ε}{\ensuremath{\varepsilon}}
\newunicodechar{λ}{\ensuremath{\lambda}}
\newunicodechar{γ}{\ensuremath{\gamma}}
\newunicodechar{δ}{\ensuremath{\delta}}
\newunicodechar{ρ}{\ensuremath{\rho}}
\newunicodechar{π}{\ensuremath{\pi}}
\newunicodechar{ω}{\ensuremath{\omega}}

\fi

\author{}
\date{}

\begin{document}

\hypertarget{jagarin-a-three-layer-architecture-for-hibernating-personal-duty-agents-on-mobile}{%
\section{Jagarin: A Three-Layer Architecture for Hibernating Personal
Duty Agents on
Mobile}\label{jagarin-a-three-layer-architecture-for-hibernating-personal-duty-agents-on-mobile}}

\textbf{Ravi Kiran Kadaboina}

\emph{Independent Researcher}

\emph{Jagarin - Sanskrit for wakeful, awake, vigilant}

\hypertarget{abstract}{%
\subsection{Abstract}\label{abstract}}

Personal AI agents face a deployment paradox on mobile: persistent
background execution drains the battery and conflicts with platform
background limits, yet purely reactive agents miss time-sensitive
obligations until the user remembers to ask. We present
\textbf{Jagarin}, a three-layer architecture that resolves this through
structured hibernation and demand-driven wake. \textbf{DAWN} (Duty-Aware
Wake Network) is an on-device scoring engine that runs on the platform's
periodic wake and combines four signals (duty-typed optimal action
windows, predicted user engagement, the cost of delay, and cross-duty
batching) with per-duty adaptive thresholds to decide whether a sleeping
agent should stay silent, nudge the user, or offer escalation.
\textbf{ARIA} (Agent Relay Identity Architecture) is a commercial email
identity proxy that turns institutional email into structured duty
records and routes messages by category, removing manual data entry.
\textbf{ACE} (Agent-Centric Exchange) is a protocol for machine-readable
communication from institutions to personal agents, intended to make
email parsing unnecessary in the long run. DAWN and ACE are specified
and evaluated in companion papers; this paper describes how the three
layers fit together and a working Flutter prototype on Android that
combines them with an ephemeral cloud agent invoked only when the user
asks. Behavioural signals, thresholds and scoring never leave the
device, and every record ARIA extracts is sealed to the device's public
key before it is stored, so the relay holds only ciphertext it cannot
read. Cloud model exposure is limited to parsing commercial email and to
user-initiated escalation, which receives only the structured duty
record.

\hypertarget{introduction}{%
\subsection{1. Introduction}\label{introduction}}

The promise of personal AI agents managing everyday obligations
(insurance renewals, prescription refills, appointment follow-ups,
subscription expirations) is well understood. The deployment reality is
not.

Existing reminder systems answer the wrong question. They ask: \emph{how
close is the deadline?} The right question is: \emph{what is the cost of
not acting right now, for this person, at this moment?} A car insurance
renewal 45 days away might be urgent today (competing quotes take time
to gather) and a poor use of attention tomorrow morning (the user has
back-to-back meetings). A prescription refill 14 days before running out
might be the worst thing to raise on a busy Monday and the best on a
Saturday when the user passes the pharmacy.

Existing personal agent platforms fit mobile poorly for four related
reasons:

\begin{enumerate}
\def\labelenumi{\arabic{enumi}.}
\item
  \textbf{Execution model mismatch.} Persistent background agents
  conflict with iOS and Android background-execution limits. Android
  WorkManager {[}1{]} and iOS BGTaskScheduler {[}2{]} provide controlled
  wake windows, but to our knowledge no prior agent framework models
  \emph{whether} action is worthwhile at a given wake.
\item
  \textbf{Cold-start dependency.} Duty-aware agents need structured
  obligation data. No user manually enters ``insurance renews October
  15, optimal action window opens September 5.'' Without automatic duty
  discovery, agents have nothing to reason over.
\item
  \textbf{Communication layer mismatch.} Institutional communications
  (receipts, renewal notices, confirmations) are designed for human
  eyes. An agent receiving an HTML email that says ``Your policy renews
  soon'' cannot reliably extract a deadline, a reference number, an
  optimal action window, or what it is allowed to do next.
\item
  \textbf{Alignment mismatch.} An agent deployed by an institution
  optimises for that institution's objectives, not the user's. An
  insurer's own app will remind you to renew with that insurer; it has
  no reason to compare competitor quotes. A retailer's rewards agent
  surfaces the redemption paths that maximise engagement with that
  retailer, not the ones with the highest value to the user. If twenty
  institutions each deploy a per-app agent, all twenty decide
  independently when to interrupt the user, with no shared interrupt
  budget. Jagarin is user-deployed; its optimisation target is user
  value.
\end{enumerate}

Jagarin addresses the four problems with one layer each, plus the
prototype that joins them. Each layer is independently deployable.

\textbf{Contributions:}

\begin{itemize}
\tightlist
\item
  \textbf{DAWN}: a multi-signal heuristic for deciding, at each platform
  wake, whether a hibernating agent should act, framed as
  opportunity-cost minimisation over duty-typed value curves (§3; full
  specification and evaluation in {[}9{]})
\item
  \textbf{ARIA}: a commercial identity proxy that turns institutional
  email into duty records and routes the commercial inbox by message
  category, so that promotions do not flood the duty registry (§4)
\item
  \textbf{ACE}: a protocol for direct institution-to-agent
  communication, filling the gap between human-targeted and
  transaction-targeted protocols (§5; full specification in {[}10{]})
\item
  \textbf{Jagarin}: a working prototype that combines the three layers
  on Android with an ephemeral cloud escalation path (§6)
\end{itemize}

All institutions named in the examples are fictional.

\hypertarget{system-architecture}{%
\subsection{2. System Architecture}\label{system-architecture}}

The three layers are vertically integrated but independently separable:

\begin{figure}
\centering
\includegraphics[width=1\textwidth,height=\textheight]{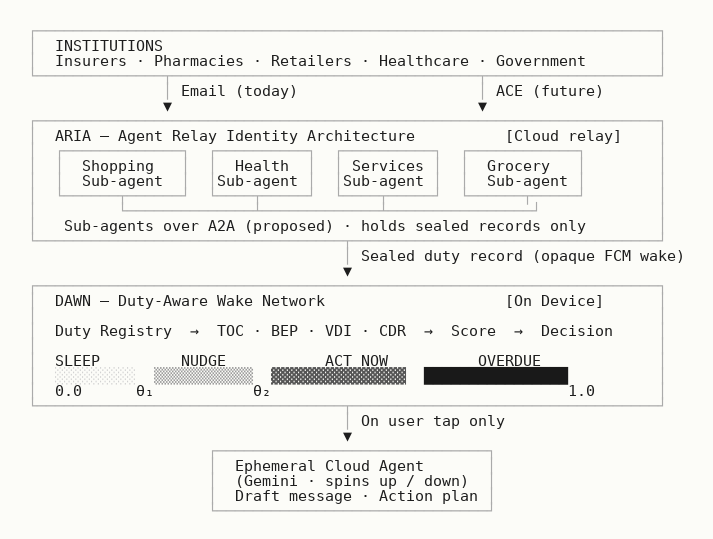}
\caption{Jagarin three-layer system architecture. Institutions feed ARIA
via email today and ACE in future; ARIA classifies messages and hands
duty records to the on-device DAWN engine via FCM; the ephemeral cloud
agent runs only on an explicit user tap. The vertical sub-agents are
proposed (§4.5); the prototype runs ARIA as a single backend. Cross-duty
resonance is computed on the device.}
\end{figure}

\textbf{Key architectural properties:}

\begin{itemize}
\tightlist
\item
  \textbf{Behaviour stays on the device.} DAWN scores, thresholds,
  interaction history and the engagement signals behind BEP are computed
  and stored only on the device, in an encrypted Hive database whose key
  is held in the platform keystore. None of it is transmitted.
\item
  \textbf{Sealed relay.} ARIA encrypts every record it extracts to the
  device's public key before storing it, so it holds only ciphertext it
  cannot open, and deletes it once the device has it (§4.4). The push
  that wakes the device carries only an opaque identifier.
\item
  \textbf{Scoped model exposure.} ARIA's Tier-2 parser sends commercial
  email to a cloud LLM (Gemini 2.5 Flash). This is a deliberate
  trade-off: the dedicated ARIA address receives only institutional
  traffic, which is less sensitive than personal correspondence. The
  escalation agent receives only the structured duty record, sent by the
  device with the request, and keeps nothing.
\item
  \textbf{No continuous background execution.} Android WorkManager wakes
  DAWN periodically (15-minute minimum) and a high-priority push wakes
  it when a duty arrives, but Android defers periodic work for rarely
  opened apps (App Standby Buckets). The deadline reminder therefore
  does not depend on those wakes: the app books each duty's
  urgency-floor reminder with the operating system's alarm manager,
  which delivers it on time when the user permits exact alarms and
  otherwise possibly later, by up to a day once a restricted app's daily
  alarm quota is used (§6.1). The periodic wakes decide the earlier
  nudges. Only Android has been tested; iOS BGTaskScheduler grants wakes
  at the system's discretion.
\end{itemize}

\hypertarget{dawn-duty-aware-wake-network}{%
\subsection{3. DAWN: Duty-Aware Wake
Network}\label{dawn-duty-aware-wake-network}}

DAWN is specified and evaluated in the companion paper {[}9{]}; this
section summarises what the architecture relies on. At each platform
wake, DAWN scores every registered duty and returns one of SLEEP, NUDGE,
ACT NOW or OVERDUE. It does not schedule wake-ups itself.

\textbf{Signals.} Each signal lies in {[}0, 1{]}. The \textbf{Temporal
Opportunity Curve (TOC)} encodes the fact that every duty type has an
optimal action window before its deadline. With \(t\) the days
remaining, TOC is an asymmetric Gaussian that peaks at a duty-type
centre \(\mu_\tau\), with width \(\sigma_R\) on the early side and
\(\sigma_L\) on the urgent side:

\[
\mathrm{TOC}(t,\tau) \;=\;
\begin{cases}
\exp\!\left( -\dfrac{(t-\mu_\tau)^2}{2\sigma_R^2} \right), & t \ge \mu_\tau \quad \text{(early side)} \\[2ex]
\exp\!\left( -\dfrac{(t-\mu_\tau)^2}{2\sigma_L^2} \right), & t < \mu_\tau \quad \text{(urgent side)}
\end{cases}
\]

An obligation 45 days out can therefore score higher than one 5 days
out, if the first is inside its window and the second has already lost
its leverage. The other three signals are:

\begin{itemize}
\tightlist
\item
  \textbf{Behavioural Engagement Predictor (BEP).} The probability that
  the user responds to a notification now, estimated on the device from
  hour of day, day of week, charging and Wi-Fi state, the recent ignore
  streak, and hours since the app was last opened.
\item
  \textbf{Value Decay Integral (VDI).} The rate at which waiting
  destroys value: zero while the user is still approaching the window,
  positive once the peak has passed. The implementation takes a one-day
  difference of TOC and clips it at zero, so VDI never lowers the
  composite.
\item
  \textbf{Cross-Duty Resonance (CDR).} A batching bonus when another
  active duty shares a counterparty, an overlapping window or an agent
  capability, so one session can handle both (``your auto and home
  insurance both renew within six weeks''). CDR is unavailable to
  per-app agents: an insurer's app sees the auto policy it issued, not
  the home policy held elsewhere. Only an agent that holds duties across
  counterparties can see the overlap.
\end{itemize}

\textbf{Composite and decisions.} The composite score is

\[
S = 0.35\,\mathrm{TOC} + 0.25\,\mathrm{BEP} + 0.25\,\mathrm{VDI} + 0.15\,\mathrm{CDR}.
\]

Two per-duty thresholds \(\theta^1 < \theta^2\) divide it into SLEEP
(\(S < \theta^1\)), NUDGE (\(\theta^1 \le S < \theta^2\)) and ACT NOW
(\(S \ge \theta^2\)). ACT NOW makes the cloud agent available;
escalation itself is always user-initiated. Within a duty-type urgency
floor before the deadline the decision is forced to ACT NOW, and past
the deadline it is OVERDUE. Duties that need the user's physical
presence, such as a buy-online-pick-up-in-store order, are capped at
NUDGE, including at the urgency floor, since no cloud agent can
substitute for being there.

\textbf{Adaptive thresholds.} Thresholds adapt per duty with a simple
step rule: each response to a notification lowers both thresholds by
\(\alpha\), and each ignored notification raises both by \(\alpha\),
with \(\alpha = 0.05\), bounds \([0.15, 0.75]\), a minimum gap of 0.05
and a ceiling of 0.55 on \(\theta^1\); initial values are
\((0.30, 0.65)\). The rule settles where the user responds to half of
the notifications that fire. This rule corrects the threshold update
given in earlier versions of this paper. Two users with identical duties
may therefore be reminded on different days. A user who answers every
nudge drives both thresholds to their floor and is nudged earlier, which
is the intent; how often they are notified is bounded separately by the
notification policy (§6), which shows at most one notification per duty
per day whatever the thresholds are.

\textbf{Evidence and limits.} BEP is calibrated in {[}9{]} on a public
dataset of 3.4 million notification events from 342 users {[}11{]}: a
logistic model reaches a test AUC of 0.69, against 0.62 for the
rule-based heuristic. The dataset is mostly social and messaging
notifications, so transfer to institutional duties is a working
hypothesis. BEP and the thresholds only shift \emph{when} within the
window a duty surfaces; the urgency floor ignores both, so neither can
suppress a reminder before the deadline. A user who keeps ignoring
nudges about a long-window duty is next notified only at the urgency
floor {[}9{]}.

\hypertarget{aria-agent-relay-identity-architecture}{%
\subsection{4. ARIA: Agent Relay Identity
Architecture}\label{aria-agent-relay-identity-architecture}}

DAWN needs structured duty data, and manual entry is impractical. ARIA
solves the cold-start problem by converting institutional email into
duties automatically.

\hypertarget{commercial-identity-separation}{%
\subsubsection{4.1 Commercial Identity
Separation}\label{commercial-identity-separation}}

ARIA gives the user a dedicated agent email address (for example
\texttt{user@aria.example}) that is handed only to commercial entities.
The user's personal inbox receives only intentional correspondence, and
the agent address receives only commercial traffic, which is what makes
cloud parsing of it an acceptable trade-off (§2).

\hypertarget{purchase-pattern-model-ppm-and-social-pattern-model-spm}{%
\subsubsection{4.2 Purchase Pattern Model (PPM) and Social Pattern Model
(SPM)}\label{purchase-pattern-model-ppm-and-social-pattern-model-spm}}

Every invoice and receipt flowing through ARIA can be parsed into a
Purchase Pattern Model: a private, on-device record of what the user
buys, from whom, how often and at what price point. Marketing email is
scored against it: an offer in a category the user bought from recently
scores high, an unsolicited brand promotion scores near zero, and offers
below threshold are archived without notifying the user. PPM is the
relevance signal the DAWN composite {[}9{]} uses for a Commercial
Opportunity:

\begin{verbatim}
PPM = 0.45·category_affinity + 0.25·counterparty_loyalty
    + 0.20·price_band_match  + 0.10·seasonal_match
\end{verbatim}

Each sub-score lies in {[}0, 1{]} and is computed from the on-device
purchase graph: recency-weighted purchases in the offer's category
(90-day decay), the share of the user's purchases in that category made
with the sender, closeness of the offer's price tier to the user's usual
tier, and a seasonal match. The Social Pattern Model is the parallel
signal for Social/Platform Updates, weighting how often the user opens
the platform (0.40), responds to that content type (0.35) and recognises
the author (0.25).

DAWN's 0.35 / 0.25 / 0.25 / 0.15 allocation is the Temporal Obligation
composite; the other categories use their own weights with these
relevance signals added:

\begin{longtable}[]{@{}lcccccc@{}}
\toprule
Category & TOC & BEP & VDI & CDR & PPM & SPM\tabularnewline
\midrule
\endhead
Temporal Obligation & 0.35 & 0.25 & 0.25 & 0.15 & n/a &
n/a\tabularnewline
Commercial Opportunity & 0.10 & 0.15 & 0.05 & 0.10 & 0.60 &
n/a\tabularnewline
Rewards (with expiry) & 0.30 & 0.10 & 0.30 & 0.10 & 0.20 &
n/a\tabularnewline
Social Update & 0.05 & 0.25 & n/a & n/a & n/a & 0.70\tabularnewline
\bottomrule
\end{longtable}

PPM, SPM and these weights are a specification: the prototype does not
yet compute them and gives every commercial message a relevance of 1.0.

\hypertarget{category-routing}{%
\subsubsection{4.3 Category Routing}\label{category-routing}}

ARIA does not register every commercial message as a duty. It routes by
category:

\begin{longtable}[]{@{}llll@{}}
\toprule
\begin{minipage}[b]{0.22\columnwidth}\raggedright
Category\strut
\end{minipage} & \begin{minipage}[b]{0.22\columnwidth}\raggedright
Examples\strut
\end{minipage} & \begin{minipage}[b]{0.22\columnwidth}\raggedright
ARIA action\strut
\end{minipage} & \begin{minipage}[b]{0.22\columnwidth}\raggedright
DAWN integration\strut
\end{minipage}\tabularnewline
\midrule
\endhead
\begin{minipage}[t]{0.22\columnwidth}\raggedright
\textbf{Temporal Obligation}\strut
\end{minipage} & \begin{minipage}[t]{0.22\columnwidth}\raggedright
Insurance renewal, subscription expiry, store pickup, return
deadline\strut
\end{minipage} & \begin{minipage}[t]{0.22\columnwidth}\raggedright
Extract deadline and optimal window, register a duty record\strut
\end{minipage} & \begin{minipage}[t]{0.22\columnwidth}\raggedright
Full TOC, VDI and CDR scoring\strut
\end{minipage}\tabularnewline
\begin{minipage}[t]{0.22\columnwidth}\raggedright
\textbf{Commercial Opportunity}\strut
\end{minipage} & \begin{minipage}[t]{0.22\columnwidth}\raggedright
Flash sales, weekend promotions, personalised offers\strut
\end{minipage} & \begin{minipage}[t]{0.22\columnwidth}\raggedright
PPM relevance score\strut
\end{minipage} & \begin{minipage}[t]{0.22\columnwidth}\raggedright
Store and notify at low priority if relevant; otherwise archive\strut
\end{minipage}\tabularnewline
\begin{minipage}[t]{0.22\columnwidth}\raggedright
\textbf{Rewards Signal}\strut
\end{minipage} & \begin{minipage}[t]{0.22\columnwidth}\raggedright
Points balance, tier upgrade, points expiry warning\strut
\end{minipage} & \begin{minipage}[t]{0.22\columnwidth}\raggedright
Check for an expiry date and redeemable value\strut
\end{minipage} & \begin{minipage}[t]{0.22\columnwidth}\raggedright
Register a duty if both are present; otherwise update the rewards
record\strut
\end{minipage}\tabularnewline
\begin{minipage}[t]{0.22\columnwidth}\raggedright
\textbf{Social/Platform Update}\strut
\end{minipage} & \begin{minipage}[t]{0.22\columnwidth}\raggedright
Follower milestones, community activity\strut
\end{minipage} & \begin{minipage}[t]{0.22\columnwidth}\raggedright
Engagement score at ingest\strut
\end{minipage} & \begin{minipage}[t]{0.22\columnwidth}\raggedright
Notify if relevant; no duty registered\strut
\end{minipage}\tabularnewline
\bottomrule
\end{longtable}

The taxonomy mirrors ACE's message categories (§5) and keeps promotional
volume out of the duty registry: an insurance renewal and a flash sale
are structurally different even when both arrive by email. ACE defines a
fifth category, Security Alert, which bypasses scoring and interrupts
the user by priority class; ARIA does not route it yet, so such an email
arrives as an ordinary item.

\hypertarget{automatic-duty-ingestion}{%
\subsubsection{4.4 Automatic Duty
Ingestion}\label{automatic-duty-ingestion}}

ARIA first classifies each message: Gemini 2.5 Flash chooses the
category, treating any email that states a deadline the recipient must
meet as an obligation even if it also advertises, with keyword rules as
the offline fallback. Obligations go to the Tier-2 parser (Gemini 2.5
Flash), with the Tier-1 keyword and regular-expression parser as the
fallback; the other categories use Tier-1 only. PDF attachments are
converted to text and appended to the parsing context. For an obligation
ARIA then:

\begin{enumerate}
\def\labelenumi{\arabic{enumi}.}
\tightlist
\item
  Extracts the counterparty, deadline, optimal action window, reference
  number, value at stake and escalation capability
\item
  Builds a DAWN duty record
\item
  Seals the record to the device's public key, stores only the
  ciphertext, and wakes the device with a Firebase Cloud Messaging data
  message that carries only an opaque identifier
\item
  The device fetches and decrypts the record, stores it in its encrypted
  registry and acknowledges it, and ARIA deletes the ciphertext;
  periodic sync is the fallback
\end{enumerate}

No user action is needed between receipt and registration. For rewards
messages received through ACE with an expiry date and a redeemable value
above USD 5.00, the backend registers a duty with a cliff at the expiry,
turning a passive balance notice into a tracked obligation.

\hypertarget{vertical-sub-agents}{%
\subsubsection{4.5 Vertical Sub-Agents}\label{vertical-sub-agents}}

The prototype uses a single FastAPI backend that performs all routing
in-process. A future version could split parsing into vertical
sub-agents (Shopping, Health, Services, Grocery) coordinated over A2A
{[}5{]}. Like the single backend, each would see an email only while
parsing it and store only the sealed record, so the sub-agents cannot
compare duties with one another; cross-duty resonance stays on the
device, the only place that can read every duty. The split is not
implemented; the part that is implemented is the category routing that
keeps promotions out of the obligation pipeline.

\hypertarget{ace-agent-centric-exchange-protocol}{%
\subsection{5. ACE: Agent-Centric Exchange
Protocol}\label{ace-agent-centric-exchange-protocol}}

ARIA parses email because institutions write for people. ACE {[}10{]}
specifies what institutions would send if they wrote for agents. Email
(RFC 5321 {[}7{]}) and iCalendar (RFC 5545 {[}8{]}) assume a human
recipient, and agentic-commerce protocols such as UCP {[}3{]} cover the
point of purchase rather than the ongoing relationship; A2A {[}5{]}, ACP
{[}4{]} and MCP {[}6{]} address agent-to-agent tasks and tool calls.

Every ACE message carries a universal envelope with four core schemas:
ACE-TEMP (deadline, urgency class and optional action-window hints,
which map directly onto DAWN's TOC parameters), ACE-VALUE (value at
stake), ACE-SCOPE (what the agent may do without asking) and ACE-TRUST
(affiliate and recommendation-basis disclosure). Messages fall into five
categories, four of which ARIA routes today (§4.3). The companion paper
gives a JSON-LD vocabulary, SHACL shapes that make message validity
machine-checkable, a TLA+ model of the duty-registration lifecycle and
SPARQL query patterns. Each message carries a sender signature, which
the receiver verifies only against a key served from the sender's own
domain, so a spoofed message is rejected before it reaches the registry.
ACE reuses A2A's Agent Card for discovery, and an ACE message can also
travel inside ordinary email as a structured part, which gives senders
an incremental adoption path {[}10{]}.

\hypertarget{implementation-jagarin}{%
\subsection{6. Implementation: Jagarin}\label{implementation-jagarin}}

We implemented the three layers in a research prototype,
\textbf{Jagarin}, demonstrated on Android (API 26+). It is written in
Flutter, but the iOS build has not been tested.

\textbf{Mobile application (Flutter/Dart):}

\begin{itemize}
\tightlist
\item
  DAWN engine in pure Dart; no network access during scoring
\item
  Background execution: WorkManager (15-minute periodic task) plus an
  FCM wake on ingestion; urgency-floor reminders booked with the alarm
  manager, one per floor day, exact when the user grants ``Alarms \&
  reminders'' access and inexact otherwise, re-booked on every change
  and after a reboot
\item
  Notification policy: a duty entering ACT NOW or OVERDUE notifies once
  and then at most once a day while it stays there; only the top NUDGE
  per wake is shown, at most once a day per duty
\item
  Adaptive thresholds persisted per duty and updated on notification
  interaction. The prototype implements the step rule and bounds of §3;
  the gap and ceiling logic, and the calibrated TOC parameters of
  {[}9{]}, exist in the reference simulation and have not yet been
  ported.
\item
  UI: duty list with DAWN decision badges and a per-duty score breakdown
\end{itemize}

\textbf{ARIA backend (Python/FastAPI):}

\begin{itemize}
\tightlist
\item
  Model-based category classifier and Tier-2 Gemini 2.5 Flash
  extraction, with keyword and regex rules as the offline fallback
\item
  PDF attachment text extraction via PyMuPDF
\item
  ACE ingest endpoint (\texttt{POST\ /ace/ingest}) accepting envelopes
  with the four core schemas; it is protected by an API key and does not
  yet verify ACE sender signatures
\item
  A2A discovery via an Agent Card at
  \texttt{GET\ /.well-known/agent-card.json}
\item
  Sealed-box encryption of every stored record to the device key
  (PyNaCl), deletion on acknowledgement, and an opaque FCM wake on
  ingestion
\end{itemize}

\textbf{Ephemeral cloud agent:}

\begin{itemize}
\tightlist
\item
  Invoked only on an explicit user tap; stateless: receives the duty
  record from the device and returns a recommendation, action points and
  a draft message
\item
  If the duty carries an institution's action endpoint (from ACE), the
  agent uses it; otherwise Gemini 2.5 Flash drafts the plan, with a
  rule-based fallback when no model is available
\end{itemize}

\textbf{Duty types implemented (12):} Insurance Renewal, Prescription
Refill, Wellness Visit, Subscription Renewal, Vehicle Service, Return
Deadline, License Renewal, Support Follow-up, Tax Deadline, Travel
Check-in, Store Pickup, Custom.

\begin{center}
\begin{minipage}[t]{0.30\textwidth}
\centering
\includegraphics[width=\linewidth]{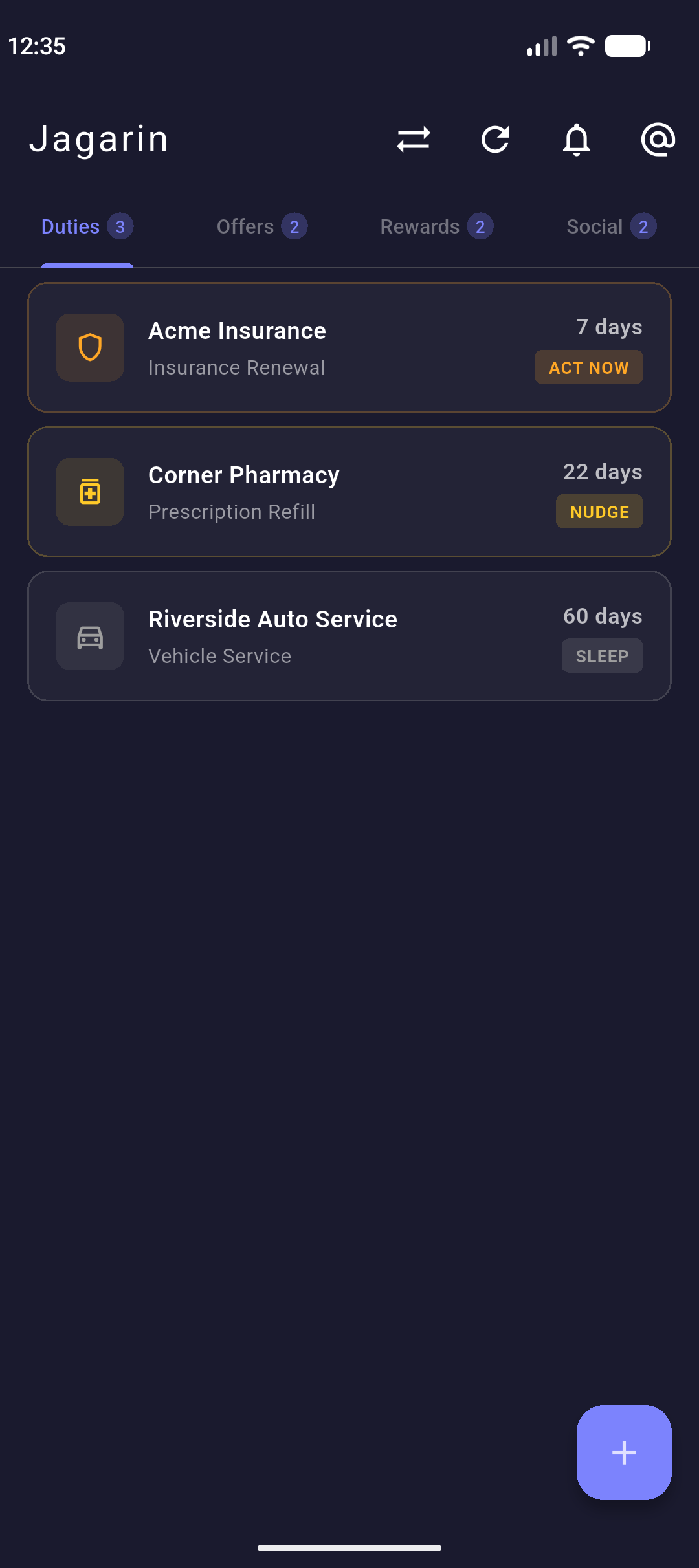}\\[0.4em]
{\small Figure 2. Duties tab. DAWN decisions appear as badges: vehicle service in 60 days (SLEEP), prescription refill in 22 days (NUDGE), insurance renewal in 7 days (ACT NOW, forced by the urgency floor).}
\end{minipage}\hfill%
\begin{minipage}[t]{0.30\textwidth}
\centering
\includegraphics[width=\linewidth]{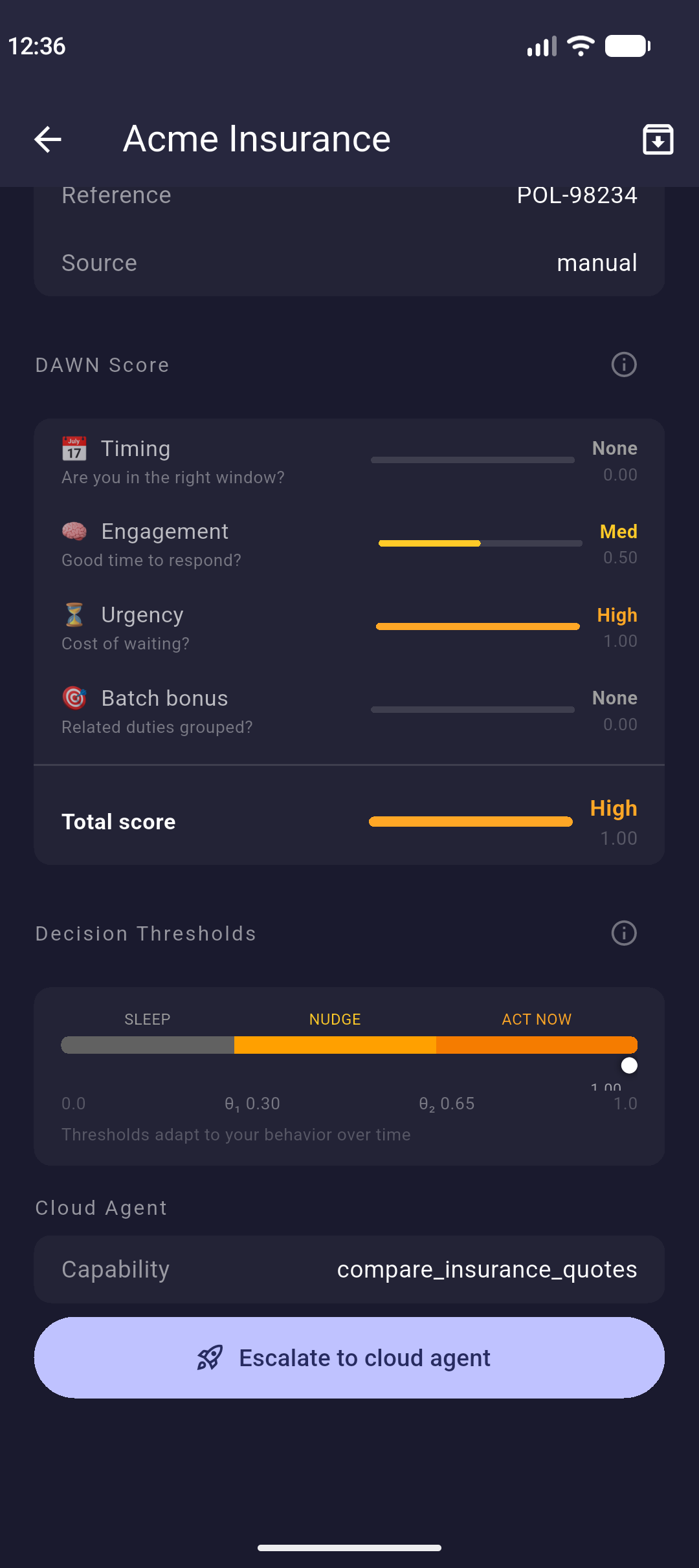}\\[0.4em]
{\small Figure 3. Duty detail: the four DAWN signals, the composite score and the duty's adaptive thresholds, with escalation offered at ACT NOW.}
\end{minipage}\hfill%
\begin{minipage}[t]{0.30\textwidth}
\centering
\includegraphics[width=\linewidth]{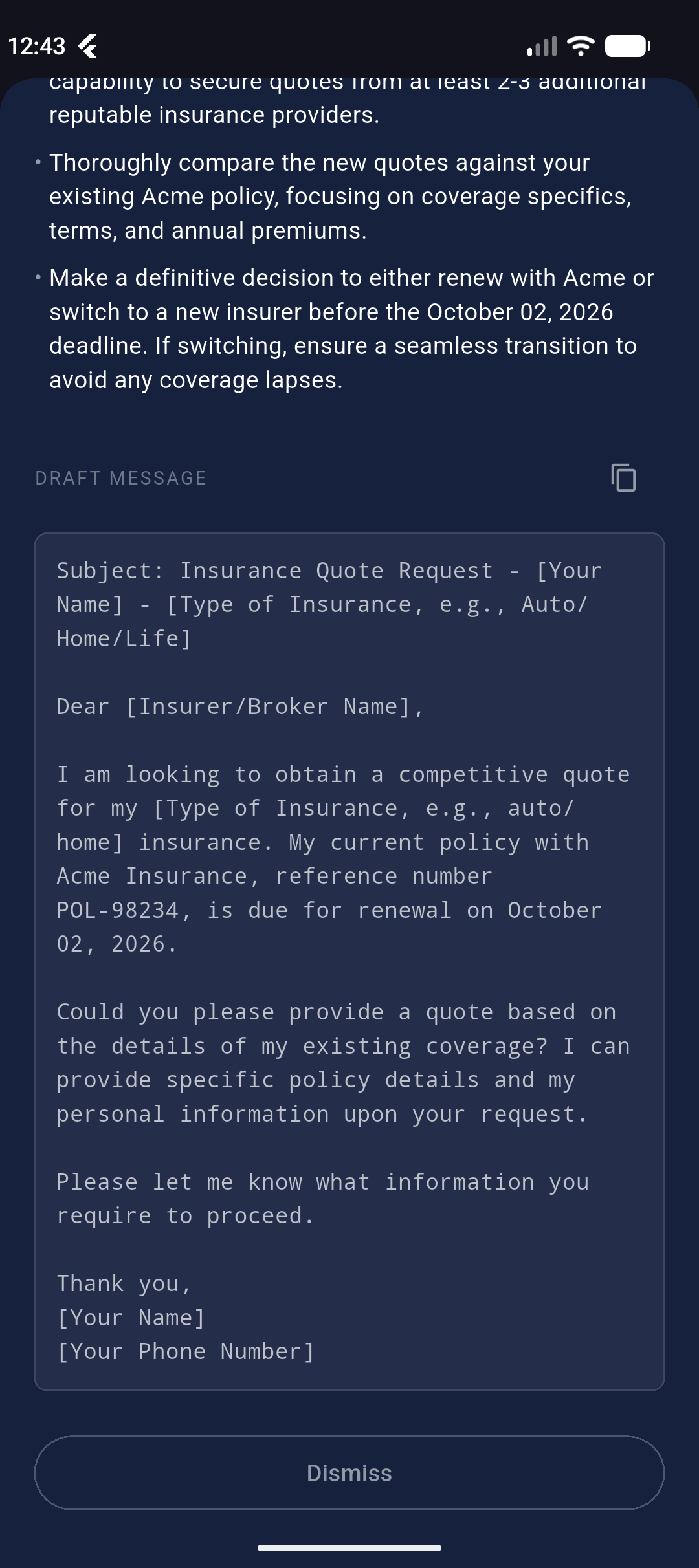}\\[0.4em]
{\small Figure 4. Cloud agent result from Gemini 2.5 Flash: the last action points and a draft quote request built only from the duty record the device sent.}
\end{minipage}
\end{center}

Screenshots show sample duties seeded with fictional institutions.

\hypertarget{deadline-reminder-delivery}{%
\subsubsection{6.1 Deadline reminder
delivery}\label{deadline-reminder-delivery}}

We checked on an Android emulator, one run per condition, that the
booked reminder is delivered when Android throttles the app: after
confirming the booking with \texttt{dumpsys\ alarm}, we moved the app to
the \texttt{restricted} App Standby bucket, turned the screen off,
forced deep idle (Doze), and read the posted notification's timestamp.

\begin{longtable}[]{@{}ll@{}}
\toprule
\begin{minipage}[b]{0.47\columnwidth}\raggedright
Reminder\strut
\end{minipage} & \begin{minipage}[b]{0.47\columnwidth}\raggedright
Delivered\strut
\end{minipage}\tabularnewline
\midrule
\endhead
\begin{minipage}[t]{0.47\columnwidth}\raggedright
Exact (``Alarms \& reminders'' granted), restricted + deep idle\strut
\end{minipage} & \begin{minipage}[t]{0.47\columnwidth}\raggedright
35 ms after the booked time\strut
\end{minipage}\tabularnewline
\begin{minipage}[t]{0.47\columnwidth}\raggedright
Inexact (not granted), restricted + deep idle\strut
\end{minipage} & \begin{minipage}[t]{0.47\columnwidth}\raggedright
about 4 minutes late, at the end of Android's batching window\strut
\end{minipage}\tabularnewline
\begin{minipage}[t]{0.47\columnwidth}\raggedright
Inexact, restricted, that day's alarm quota already used\strut
\end{minipage} & \begin{minipage}[t]{0.47\columnwidth}\raggedright
deferred by App Standby to about 23 hours later (read from
\texttt{dumpsys}, not waited out)\strut
\end{minipage}\tabularnewline
\bottomrule
\end{longtable}

With the user's permission the reminder arrived on time in the
restricted bucket under deep idle; without it, Android may shift it
within its batching window or, once a restricted app's daily alarm quota
is used, to the next day, so the app asks for the permission once at
setup. A user force-stop cancels any app's alarms until the app is
opened again. Battery Saver modes and manufacturers' own battery
managers, which can go beyond Android's rules, were not tested. The
procedure and raw \texttt{dumpsys} excerpts are in the research
repository (\texttt{eval/floor\_reminder/}).

\hypertarget{discussion}{%
\subsection{7. Discussion}\label{discussion}}

\textbf{User alignment as a design property.} Jagarin is user-deployed
rather than institution-deployed, and this is a structural choice. An
institution-deployed agent's hint for an insurance renewal will be
``renew with us''; Jagarin's is ``compare the renewal quote against
other insurers''. An institution's notification logic predicts when
engagement serves its goals; DAWN's BEP predicts when the user is
receptive. The mechanism that keeps Jagarin aligned is architectural
rather than a policy: behavioural data never reaches an institution, no
duty state is exposed to counterparties, and the escalation capability
comes from the user or from ACE messages the user receives, not from the
institution whose product is being evaluated.

\textbf{Privacy and threat model.} What the user does (app opens, which
nudges they answer, the thresholds learned from that, and every score)
is computed and stored only on the device and never transmitted. What
institutions send reaches the device through ARIA as a sealed relay:
records are encrypted to the device's public key before storage, the
push service sees only an opaque identifier, and the ciphertext is
deleted once delivered, so a compromised relay holds nothing readable.
ARIA does read each commercial email in memory while parsing it, and the
Tier-2 parser sends it to a cloud LLM; we accept this because that
address receives only institutional traffic the institution already
holds, and an on-device parser would close it. Because anyone can send
mail to the ARIA address, extracted fields are treated as untrusted data
and nothing is sent to an institution except on an explicit user tap.

\textbf{Platform compliance.} DAWN makes no network calls during a wake
cycle, and cloud invocation requires a user action, which keeps the
design within App Store and Play Store background rules. This has been
exercised on Android only.

\textbf{Relation to agent email infrastructure.} Since the first version
of this paper, email for agents has become a product category: AgentMail
{[}15{]} and Cloudflare's Email Service {[}16{]} give agents their own
programmable inboxes. They supply the first part of ARIA (a dedicated
address with inbound hooks) as a service, and an ARIA deployment could
use either as its mail transport. They do not decide when a user should
be interrupted about what arrives, which is DAWN's role, and what they
carry is still human-oriented email rather than a structured
institutional format, which is ACE's.

\textbf{Relation to agent frameworks.} AutoGPT {[}12{]}, LangGraph
{[}13{]} and CrewAI {[}14{]} assume persistent execution with continuous
model access, which does not fit mobile background limits; any of them
could back Jagarin's escalation step. Jagarin's contribution is the
decision about \emph{when} running any agent is worthwhile, not the
agent itself.

\textbf{Limitations.} The prototype uses the rule-based BEP; the
calibrated model of {[}9{]} is not deployed. ARIA's classification and
extraction accuracy have not been measured on a representative corpus.
PPM, SPM, the non-obligation weights, Security Alert routing and ACE
signature verification are specified but not implemented, and the iOS
build is untested. We report no battery measurements. The combined
system has not been tested with real users on long-horizon duties, and
ACE depends on institutional incentives that have not yet materialised.

\hypertarget{conclusion}{%
\subsection{8. Conclusion}\label{conclusion}}

Jagarin shows how a personal duty agent can work within mobile platform
limits: it hibernates between platform wakes, keeps duty state and
behavioural signals on the device, and uses cloud models only to parse
commercial email and, when the user asks, to draft an action plan. DAWN
decides when a duty deserves the user's attention, ARIA turns today's
institutional email into duties, and ACE describes the structured
channel that would make that parsing unnecessary. DAWN and ACE are
specified and evaluated in {[}9{]} and {[}10{]}. The main open work for
the architecture as a whole is measuring ARIA's extraction on a
multi-user corpus and testing the combined system with real users over
long-horizon duties.

The DAWN and ACE research artifacts (simulation, calibration, TLA+
model, SHACL shapes and tests) are available at
\url{https://github.com/ravikiran438/jagarin-research}.

This work was done independently of the author's employment and is not
affiliated with or endorsed by the author's employer or any organisation
named in it.

\hypertarget{references}{%
\subsection{References}\label{references}}

{[}1{]} Android Developers. \emph{WorkManager}.
\url{https://developer.android.com/topic/libraries/architecture/workmanager}

{[}2{]} Apple Developer. \emph{Background Tasks}.
\url{https://developer.apple.com/documentation/backgroundtasks}

{[}3{]} UCP Authors (Google, Shopify, et al.). \emph{Universal Commerce
Protocol (UCP) Specification}, version 2026-08-25. \url{https://ucp.dev}

{[}4{]} IBM Research / BeeAI Platform. \emph{Agent Communication
Protocol (ACP)}. Linux Foundation, 2025; subsequently aligned with A2A
under LF AI \& Data. Not to be confused with OpenAI and Stripe's
\emph{Agentic Commerce Protocol}. \url{https://github.com/i-am-bee/acp}

{[}5{]} Google. \emph{Agent2Agent Protocol (A2A)}.
\url{https://a2a-protocol.org}, 2025.

{[}6{]} Anthropic. \emph{Model Context Protocol (MCP)}.
\url{https://modelcontextprotocol.io}, 2024.

{[}7{]} Klensin, J. \emph{Simple Mail Transfer Protocol}. RFC 5321,
IETF, 2008.

{[}8{]} Desruisseaux, B. (ed.). \emph{Internet Calendaring and
Scheduling Core Object Specification (iCalendar)}. RFC 5545, IETF, 2009.

{[}9{]} Kadaboina, R. \emph{DAWN: Duty-Aware Wake Network for
Hibernating Personal Agents on Device}. Zenodo, 2026.
doi:10.5281/zenodo.22941511

{[}10{]} Kadaboina, R. \emph{ACE: An Agent-Centric Exchange Protocol for
Institutional Communication to Personal AI Agents}. Zenodo, 2026.
doi:10.5281/zenodo.22941595

{[}11{]} Pielot, M., Cardoso, B., Katevas, K., Serrà, J., Matic, A.,
Oliver, N. Beyond Interruptibility: Predicting Opportune Moments to
Engage Mobile Phone Users. \emph{Proc. ACM IMWUT} 1(3), Article 91,
2017.

{[}12{]} Significant Gravitas. \emph{AutoGPT}.
\url{https://github.com/Significant-Gravitas/AutoGPT}, 2023.

{[}13{]} LangChain. \emph{LangGraph}.
\url{https://github.com/langchain-ai/langgraph}, 2024.

{[}14{]} CrewAI. \emph{CrewAI}.
\url{https://github.com/crewAIInc/crewAI}, 2024.

{[}15{]} AgentMail. \emph{Email inbox API for AI agents}.
\url{https://www.agentmail.to}

{[}16{]} Cloudflare. \emph{Email for agents: Cloudflare Email Service
now in public beta}. Cloudflare Blog, April 2026.
\url{https://blog.cloudflare.com/email-for-agents/}

\end{document}